\documentclass[11pt]{article}

\usepackage[preprint]{acl}

\usepackage{times}
\usepackage{latexsym}

\usepackage[T1]{fontenc}

\usepackage[utf8]{inputenc}

\usepackage{microtype}

\usepackage{inconsolata}

\usepackage{graphicx}
\usepackage{amsmath}
\usepackage{amssymb}
\usepackage{mathtools}
\usepackage{amsthm}
\usepackage{bbm}
\usepackage{algorithm}
\usepackage{algorithmic}  
\usepackage{enumitem}
\usepackage{multirow}
\usepackage{booktabs}
\usepackage[dvipsnames,table,xcdraw]{xcolor}
\usepackage[most]{tcolorbox}
\usepackage{fontawesome}
\usepackage[normalem]{ulem}
\useunder{\uline}{\ul}{}
\usepackage{subcaption}
\usepackage{wrapfig}
\usepackage{xurl}
\usepackage{hyperref}

\title{LongTraceRL: Learning Long-Context Reasoning from Search Agent Trajectories with Rubric Rewards}

\author{
  Nianyi Lin\thanks{Equal contribution.~Work done when interned at Zhipu.},
  Jiajie Zhang\footnotemark[1],
  Lei Hou,
  Juanzi Li \\
  Tsinghua University
}

\begin{document}
\maketitle

\begin{abstract}
Long-context reasoning remains a central challenge for large language models, which often fail to locate and integrate key information in extensive distracting content.
Reinforcement learning with verifiable rewards (RLVR) has shown promise for this task, yet existing methods are limited by low-confusability distractors and sparse, outcome-only reward signals that cannot supervise intermediate reasoning steps.
To address these issues, we introduce \textsc{LongTraceRL}.
For data construction, we generate multi-hop questions via knowledge graph random walks and leverage search agent trajectories to build \emph{tiered distractors}: documents the agent read but did not cite (high confusability) and documents that appeared in search results but were never opened (low confusability), producing training contexts that are far more challenging than those built by random sampling or one-shot search.
For reward design, we propose a \emph{rubric reward} that uses the gold entities along each reasoning chain as fine-grained, entity-level process supervision. This rubric reward is applied only to responses with correct final answers (positive-only strategy), distinguishing the reasoning quality among correct responses and preventing reward hacking.
Experiments on three reasoning LLMs (4B--30B) across five long-context benchmarks demonstrate that \textsc{LongTraceRL} consistently outperforms strong baselines and encourages comprehensive, evidence-grounded reasoning. 
Codes, datasets and models are available at
\href{https://github.com/THU-KEG/LongTraceRL}{https://github.com/THU-KEG/LongTraceRL}.
\end{abstract}

\section{Introduction}

\begin{figure}[!t]
    \centering
    \includegraphics[width=0.47\textwidth]{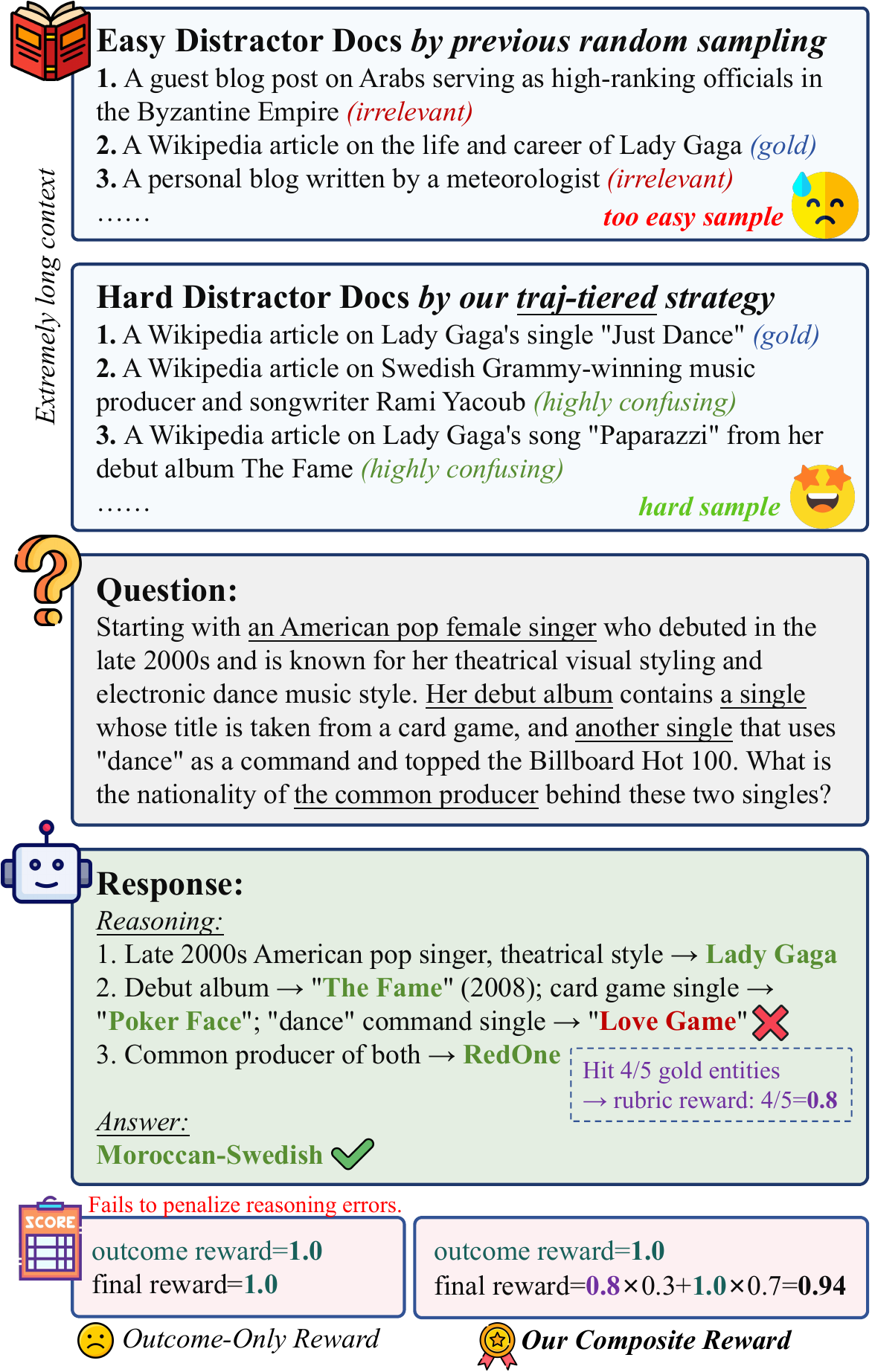}
    \caption{Comparison between prior long-context RL approaches based on easy distractors and outcome-only rewards, and our proposed \textsc{LongTraceRL}.}
    \label{fig:fail_case}
    \vspace{-1em}
\end{figure}

Long-context reasoning is a critical capability for large language models (LLMs), driving advances in both single-pass reasoning~\citep{longbench2,aa_lcr} and multi-turn autonomous agent systems~\citep{yao2023react,shinn2023reflexion,yang2024sweagent} by enabling models to extract key information across a global context, perform multi-hop inference, and stay coherent over extended text.
Despite its importance, current LLMs still struggle with long-context understanding~\citep{eapo}, especially in realistic scenarios full of distracting information. As the context length grows, they often exhibit typical failure patterns: giving hallucinated answers, relying on fragmented retrieval, or citing irrelevant passages. These limitations make long-text reasoning a major bottleneck for deploying reasoning-oriented models in real-world applications.
Recently, reinforcement learning with verifiable rewards (RLVR) has proven effective for multiple tasks such as mathematical reasoning~\citep{deepseek_r1,shao2024deepseekmath} and long-context question answering~\citep{longrlvr,loongrl}. However, current long-context RL methods have two key limitations. 
First, the quality of training data remains limited. Existing methods construct questions with few reasoning hops and shallow chains, with their distractors mostly sampled randomly from unrelated documents~\citep{loongrl,eapo}, lacking semantic relevance to the query and providing limited confusability as shown in Figure~\ref{fig:fail_case}. 
Second, the reward signal is too sparse. Existing methods primarily rely on outcome-based rewards, providing optimization guidance solely based on the correctness of the final answer. When the input spans tens or even hundreds of thousands of tokens, such rewards become very sparse and can be noisy: the model may reach the correct answer through a wrong reasoning path by chance. For example, as shown in Figure~\ref{fig:fail_case}, the model correctly answers ``Moroccan-Swedish'' while actually citing the wrong entity ``Love Game'' instead of ``Just Dance'' at an intermediate hop. Such coincidental successes satisfy the binary outcome reward but mask retrieval failures in intermediate steps.

To address these limitations, we propose \textbf{\textsc{LongTraceRL}}, which tackles both data construction and reward design. On the data side, inspired by \citet{lu2025deepdive}, we generate complex multi-hop questions with extremely long reasoning chains via knowledge graph random walks over the KILT Wikipedia snapshot~\citep{petroni2021kilt}, and introduce a novel approach that constructs distractors based on real search trajectories from a search agent. Specifically, documents that the agent read but did not cite in the final response serve as high-confusability distractors (Tier-1), while documents that appeared in search results but were never opened serve as low-confusability distractors (Tier-2). Compared to random sampling or single direct search, these distractors are more relevant to the query, forcing the model to distinguish more carefully and reason more deeply. On the reward side, we design a rubric reward that uses the gold entities at each hop of the reasoning chain as fine-grained, entity-level process supervision. This reward is applied only to responses with correct final answers (positive-only strategy), helping distinguish the reasoning quality among correct responses and effectively preventing the model from gaming the reward by skipping intermediate reasoning steps and guessing the answer directly.
Experiments on three reasoning LLMs (4B--30B) across five long-context benchmarks demonstrate that \textsc{LongTraceRL} consistently outperforms all baselines and encourages comprehensive, evidence-grounded reasoning, with Qwen3-4B achieving an average gain of 5.7 points over the base model and surpassing the strongest baseline by 2.5 points.

Our main contributions are as follows: (1) We propose a long-context training data construction method based on search agent trajectories, using tiered distractors to significantly improve the challenge and realism of the training data. (2) We design an entity-level rubric reward that provides finer-grained supervision on intermediate reasoning than the existing approaches for long-context reinforcement learning. (3) Through comprehensive experiments, we demonstrate the consistent improvements of \textsc{LongTraceRL} across multiple model families and scales.
\section{Related Work}

\begin{figure*}[!t]
    \centering    
    \includegraphics[width=\linewidth]{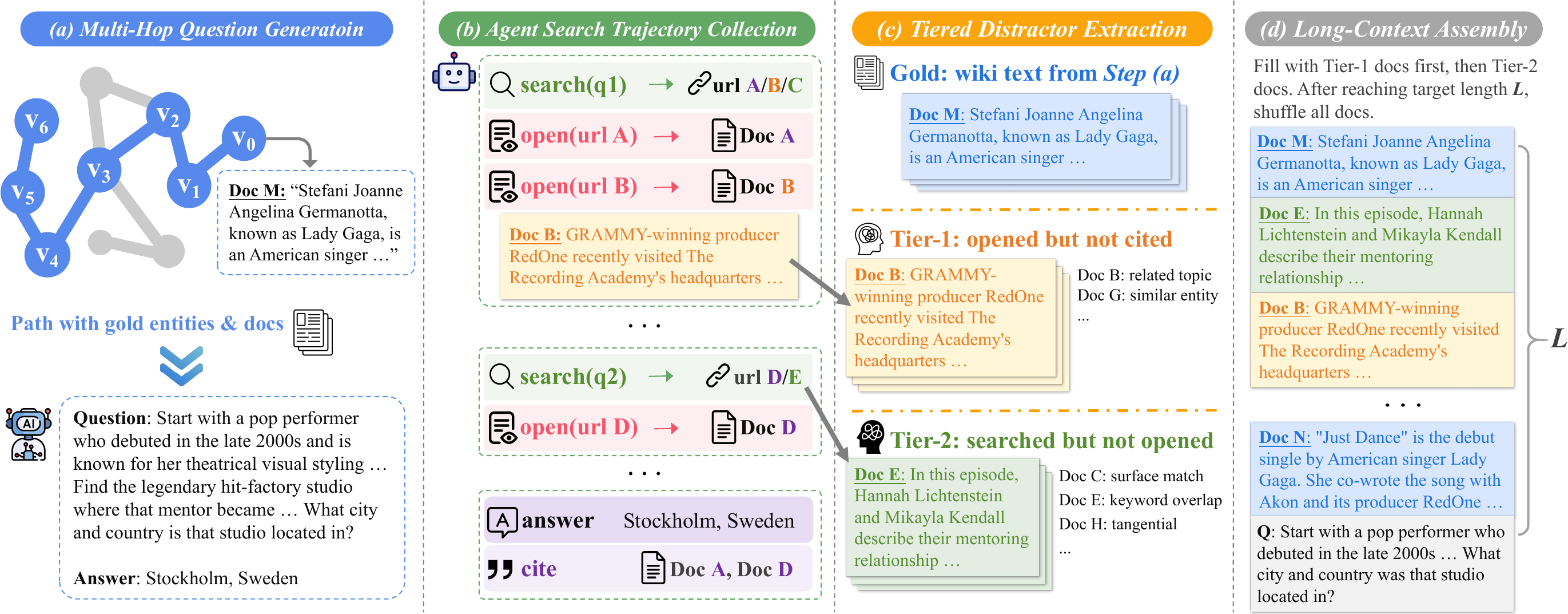}
    \caption{Overview of the \textsc{LongTraceRL} training data construction pipeline.}
    \label{fig:pipeline}
\end{figure*}

\paragraph{Long-Context Synthetic Data.}
Synthesizing long-context training data typically involves two design choices: how to construct questions and how to assemble the context.
For question construction, some work~\citep{loongrl,sealong,longreps,eapo} reuses short-context multi-hop QA datasets such as MuSiQue~\citep{trivedi2022musique} and HotpotQA~\citep{yang2018hotpotqa}, while others generate questions from scratch~\citep{longalign,longmit}. More recent work uses structured knowledge to produce deeper reasoning chains, e.g., QwenLong-L1.5~\citep{qwenlong_l1_5} samples multi-hop paths from document-derived knowledge graphs, and DeepDive~\citep{lu2025deepdive} performs random walks over the Wikipedia knowledge graph.
For context assembly, some methods use a single long document~\citep{longalign,longmit,longrlvr}, while others extend short-context QA datasets by adding distractor documents~\citep{loongrl,sealong,longreps,eapo}. These distractors are usually sampled at random and are easy to filter out. NExtLong~\citep{nextlong} improves this by using hard negative mining from dense retrieval, but its distractors are still based on embedding similarity rather than realistic search behavior, leaving a gap with practical retrieval scenarios.

\paragraph{Long-context Reinforcement Learning.}
While RLVR has proven effective on self-contained reasoning tasks such as mathematics~\citep{deepseek_r1,shao2024deepseekmath}, its adaptation to long-context scenarios remains limited, since outcome-based rewards supervise only the final answer and give no signal for intermediate reasoning over a large input. 
To address this, \citet{longrlvr} uses a chunk-level context reward based on $\text{F}_\beta$ scores between predicted and gold document chunks, \citet{eapo} provides dense process-level supervision on evidence extraction quality via a co-evolving reward model, and LongR~\citep{longr} measures relative information gain from retrieved documents under a frozen verifier.
In parallel, fine-grained process rewards have also been explored in agentic RL: \citet{zhang2026chaining} decompose deep-search answers into citation-aware rubric items for verifiable reward, and \citet{singh2025fathom} shape the reward by classifying the cognitive behavior and utility of each tool call to reduce reward hacking.
However, these methods either operate at the chunk, document, or tool-call level, or require an auxiliary LLM for evidence scoring at additional cost, leaving finer-grained, entity-level reasoning supervision unexplored.

\section{Method}
\label{sec:method}

\textsc{LongTraceRL} framework consists of two main components: (1) a data construction pipeline that synthesizes long-context training data with agent-derived distractors (\S\ref{sec:data_construction}), and (2) a reinforcement learning framework that combines outcome-based and process-based rewards (\S\ref{sec:rl_training}).

\subsection{Data Construction Pipeline}
\label{sec:data_construction}

We construct long-context training data through a four-step pipeline. The key idea is to leverage knowledge graph structure to generate multi-hop questions with verifiable reasoning chains, and then use agent search behavior to produce realistic distractors with high confusability. Figure~\ref{fig:pipeline} provides an overview of the entire pipeline.

\subsubsection{Multi-Hop Question Generation}
\label{sec:question_generation}

Inspired by \citet{lu2025deepdive}, we automatically generate multi-hop questions from the KILT Wikipedia snapshot~\citep{petroni2021kilt} through a two-stage process: knowledge graph random walk and question synthesis.

\paragraph{Knowledge Graph Random Walk.}
We perform controlled random walks over the Wikipedia hyperlink graph to collect multi-hop entity paths. Starting from a seed entity $v_0$, we walk $k(= 8)$ steps following hyperlinks to form a path $P = [v_0, v_1, \ldots, v_k]$. At each step, an LLM selects the next most relevant entity from up to five unvisited candidates. Between path collections, we insert periodic \emph{mad walks} (a few random jumps) to diversify the explored graph regions.

\paragraph{Question Synthesis.}
Given a path $P$ and the Wikipedia text of each entity, we prompt a powerful LLM (such as GPT-5.2) to generate a multi-hop question whose answer is a specific attribute of the last entity $v_k$. As shown in Figure~\ref{fig:prompt_qa_generate}, the prompt enforces several constraints: (i) the question must require step-by-step reasoning through all entities in the path, with no shortcuts; (ii) all identifying information (names, dates, locations) must be paraphrased so that the answer cannot be found by simple keyword matching; and (iii) the answer must be unique and exactly match the selected attribute. The LLM also outputs the list of intermediate gold entities along the reasoning chain. For each question, we thus obtain the question text, the ground-truth answer, and the set of gold entities $\mathcal{E} = \{e_1, e_2, \ldots, e_k\}$ with their corresponding Wikipedia passages.

\subsubsection{Agent Search Trajectory Collection}
\label{sec:trajectory_collection}

To construct realistic distractors, we leverage the behavioral trajectories of a search agent attempting to answer each generated question. We deploy an agent equipped with deep search capabilities, including issuing search queries (\textsc{search}), opening and reading retrieved documents (\textsc{open}), and citing information in its response (\textsc{cite}). We record the agent's complete search trajectory $\tau = [(a_1, d_1), (a_2, d_2), \ldots]$, where $a_t$ denotes the action type and $d_t$ the associated document.

\paragraph{Trajectory Filtering.}
To obtain correct reliable trajectories, we sample $K(=5)$ independent trajectories per question and retain only those where the agent reaches the final correct answer. For each question, one of the correct trajectories is selected for subsequent distractor extraction, and the questions where all $K$ attempts fail are discarded. This filtering ensures that the retained trajectories reflect real, meaningful, and goal-directed search behavior rather than random exploration or hallucination.

\subsubsection{Tiered Distractor Extraction}
\label{sec:distractor_extraction}

From the recorded search trajectories, we divide the retrieved documents (excluding gold evidence passages) into two tiers:
(1) \textbf{Tier-1 distractors} (high confusability): documents that the agent opened and read but did not cite in its final response, which are topically relevant and were initially deemed worth reading, making them strong distractors.
(2) \textbf{Tier-2 distractors} (low confusability): documents that appeared in search results but were never opened, which are only superficially related to the query and are less likely to mislead a careful reader.

\subsubsection{Long-Context Assembly}
\label{sec:context_assembly}

The final long-context input is assembled following a strategy named \textbf{\texttt{traj-tiered}} (short for trajectory-tiered). 
Starting from the gold passages, we first add Tier-1 distractors $\mathcal{D}_1$, which are more confusing and thus more valuable for training. If the context has not yet reached the target length $L$ after exhausting all Tier-1 distractors, we continue to fill with Tier-2 distractors $\mathcal{D}_2$. This prioritization ensures that the model is exposed to as many challenging distractors as possible. All documents are then shuffled to prevent positional shortcuts.

\subsection{RL with Rubric Reward}
\label{sec:rl_training}

Since the data construction pipeline provides gold entities along each reasoning chain, we can leverage them as process supervision during reinforcement learning.
We adopt Group Relative Policy Optimization (GRPO;~\citealp{shao2024deepseekmath}) as our RL algorithm and design a composite reward that combines an outcome reward for answer correctness with a rubric reward for reasoning quality.

\paragraph{Outcome Reward.} The binary outcome reward
$r_{\text{oc}} \in \{0, 1\}$ evaluates whether the model's final short answer is correct, as determined by an LLM judge.

\paragraph{Rubric Reward.}
The raw rubric score $\hat{r}_{\text{rb}}$ measures the recall of gold entities $\mathcal{E}$ in the model's response:
\begin{equation}
    \hat{r}_{\text{rb}} = \frac{|\{e \in \mathcal{E} \mid e \text{ appears in the response}\}|}{|\mathcal{E}|}
    \label{eq:rubric_raw}
\end{equation}
A higher $\hat{r}_{\text{rb}}$ indicates that the model referenced more of the gold entities in reasoning.

\paragraph{Group-Level Rubric Normalization.}
In GRPO, each training question is answered by a group of $G$ sampled responses. Since different questions involve different numbers of gold entities and varying difficulty levels, the raw rubric scores may span different ranges across questions. To ensure comparability, we normalize $\hat{r}_{\text{rb}}$ within each group by dividing by the group maximum when it is positive:
\begin{equation}
    r_{\text{rb}} = \begin{cases}
        \dfrac{\hat{r}_{\text{rb}}}{\max_{j \in [G]} \hat{r}_{\text{rb}}^{(j)}}, & \text{if } \max_{j \in [G]} \hat{r}_{\text{rb}}^{(j)} > 0 \\[6pt]
        0, & \text{otherwise}.
    \end{cases}
    \label{eq:rubric_norm}
\end{equation}
This rescales the rubric reward to $[0, 1]$ within each group, providing consistent process signals regardless of question difficulty.

\paragraph{Positive-Only Reward Combination.}
Since the rubric reward is based on entity recall, applying it to all responses risks \emph{reward hacking}: the model could learn to enumerate entities mentioned in the retrieved passages rather than genuinely reasoning over them, inflating the rubric score without actually solving the problem. To prevent this, we adopt a \textbf{\texttt{positive-only}} strategy in which the rubric reward is only granted to responses whose final answer is correct:
\begin{equation}
    r = \begin{cases}
        (1 - \alpha) \cdot r_{\text{oc}} + \alpha \cdot r_{\text{rb}}, & \text{if } r_{\text{oc}} > 0 \\
        0, & \text{otherwise}
    \end{cases}
    \label{eq:reward}
\end{equation}
In this formulation, the rubric reward serves to differentiate among correct responses, assigning higher scores to those that provide sound intermediate reasoning and lower scores to those that arrive at the right answer through shortcuts. Incorrect responses simply receive zero reward. The hyperparameter $\alpha \in [0, 1]$ controls the weight of process supervision: when $\alpha = 0$, the reward reduces to standard outcome-based GRPO; as $\alpha$ increases, the model receives stronger incentives to ground its reasoning in the relevant evidence passages.

\section{Experiments}
\label{sec:experiment}

\begin{table*}[!t]
\centering
\resizebox{0.9\textwidth}{!}{
\begin{tabular}{lcccccc}
\toprule
\textbf{Method} & \textbf{AA-LCR} & \textbf{MRCR} & \textbf{FRAMES} & \textbf{LongBench V2} & \textbf{LongReason} & \textbf{Avg} \\
\midrule
\multicolumn{7}{l}{\textit{DeepSeek-R1-0528-Qwen3-8B}} \\
\midrule
Base & \underline{13.8} & \textbf{27.0} & 73.2 & 25.4 & 74.1 & 42.7 \\
DocQA & 9.5 & 20.2 & \underline{73.4} & 25.9 & 73.7 & 40.6 \\
LoongRL & 10.2 & 19.0 & 72.6 & 25.5 & 73.3 & 40.1 \\
LongRLVR & 12.2 & 21.9 & 70.3 & \underline{26.8} & 73.2 & 40.9 \\
\textsc{LongTraceRL}-GRPO & \textbf{15.0} & 25.2 & 73.1 & 25.6 & \textbf{75.4} & \underline{42.9} \\
\textsc{LongTraceRL} & \textbf{15.0} & \textbf{27.0} & \textbf{74.3} & \textbf{27.5} & \underline{75.2} & \textbf{43.8} \\
\midrule
\multicolumn{7}{l}{\textit{Qwen3-4B-Thinking-2507}} \\
\midrule
Base & 33.2 & 36.2 & 76.7 & 41.7 & 78.5 & 53.3 \\
DocQA & 28.8 & \underline{41.9} & 78.3 & \textbf{44.6} & 79.9 & 54.7 \\
LoongRL & 32.0 & 38.2 & 75.8 & 41.8 & 78.7 & 53.3 \\
LongRLVR & \underline{37.5} & 41.8 & \underline{78.5} & 43.8 & \underline{80.7} & \underline{56.5} \\
\textsc{LongTraceRL}-GRPO & 34.0 & 38.9 & 76.1 & 40.7 & 78.7 & 53.7 \\
\textsc{LongTraceRL} & \textbf{41.8} & \textbf{45.8} & \textbf{79.5} & \underline{44.1} & \textbf{83.8} & \textbf{59.0} \\
\midrule
\multicolumn{7}{l}{\textit{Qwen3-30B-A3B-Thinking-2507}} \\
\midrule
Base & 47.0 & 40.8 & 80.7 & 49.9 & 84.2 & 60.5 \\
DocQA & \underline{50.2} & \underline{44.8} & \textbf{83.5} & 51.6 & \underline{86.4} & \underline{63.3} \\
LoongRL & \textbf{53.5} & 43.8 & \underline{81.9} & \textbf{52.3} & 84.9 & 63.3 \\
LongRLVR & 48.5 & 43.3 & 81.7 & 50.3 & 84.2 & 61.6 \\
\textsc{LongTraceRL}-GRPO & 48.2 & 44.7 & 79.6 & \underline{51.8} & \textbf{86.9} & 62.3 \\
\textsc{LongTraceRL} & \textbf{53.5} & \textbf{46.5} & \underline{81.9} & 51.3 & 85.4 & \textbf{63.7} \\
\bottomrule
\end{tabular}
}
\caption{Main results on long-context reasoning benchmarks.}
\label{tab:scores}
\end{table*}

\begin{figure*}[!t]
    \centering
    \begin{subfigure}{0.24\linewidth}
        \centering
        \includegraphics[width=\linewidth]{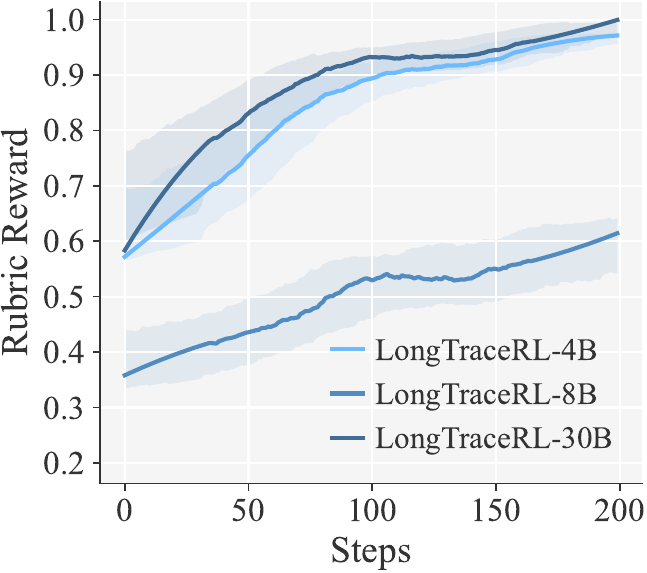}
        \caption{}
        \label{fig:rubric_reward}
    \end{subfigure}
    \begin{subfigure}{0.24\linewidth}
        \centering
        \includegraphics[width=\linewidth]{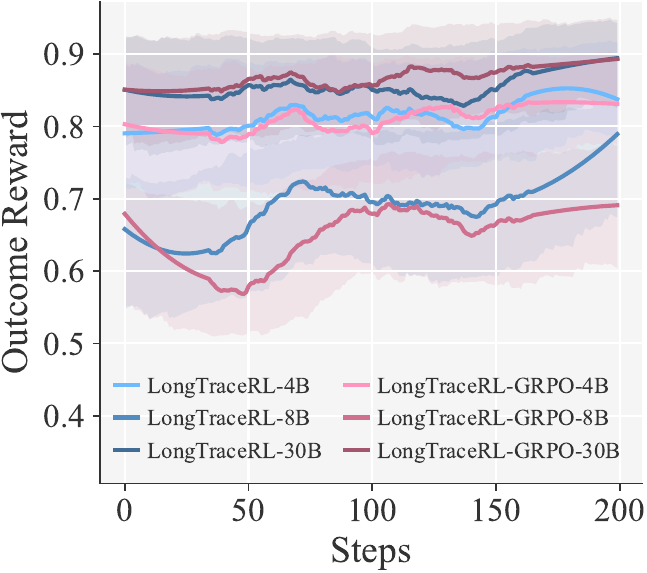}
        \caption{}
        \label{fig:outcome_reward}
    \end{subfigure}
    \begin{subfigure}{0.24\linewidth}
        \includegraphics[width=\linewidth]{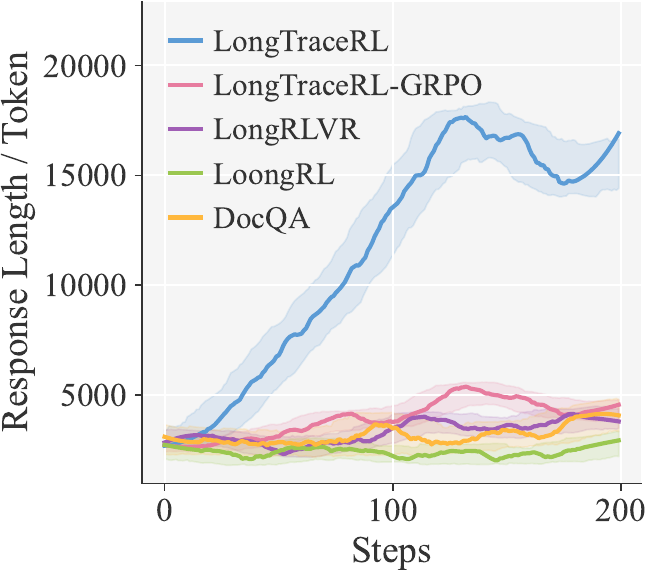}
        \caption{}
        \label{fig:resp_len}
    \end{subfigure}
    \begin{subfigure}{0.24\linewidth}
        \includegraphics[width=\linewidth]{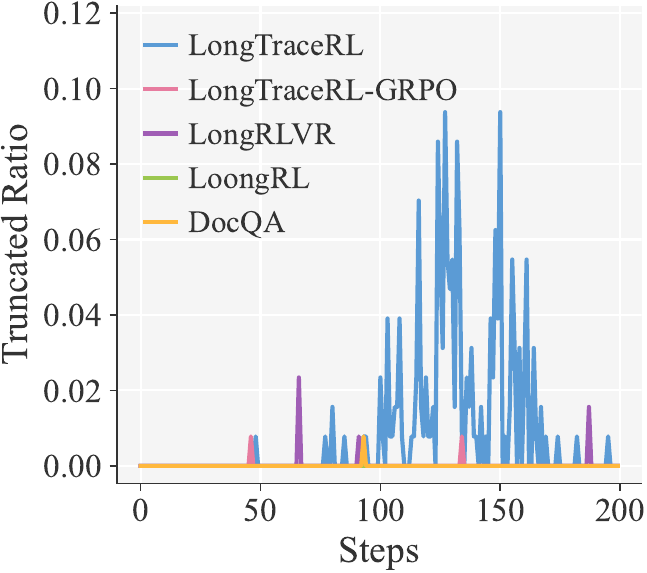}
        \caption{}
        \label{fig:truncated}
    \end{subfigure}
    \caption{From left to right: rubric and outcome reward dynamics at different scales, rollout response length and truncation rate dynamics across methods. 
    }
    \label{fig:main}
\end{figure*}

\subsection{Setup}

\begin{table*}[t]
\centering
\resizebox{0.9\textwidth}{!}{
\begin{tabular}{l|cccccc}
\toprule
\textbf{Method} & \textbf{AA-LCR} & \textbf{MRCR} & \textbf{FRAMES} & \textbf{LongBench V2} & \textbf{LongReason} & \textbf{Avg} \\
\midrule
\textit{Qwen3-4B-Thinking-2507} & 33.2 & 36.2 & 76.7 & 41.7 & 78.5 & 53.3 \\
+ \textsc{LongTraceRL}($\alpha=0.1$) & \underline{39.2} & \textbf{46.1} & 79.0 & \textbf{44.2} & \underline{82.8} & \underline{58.3} \\
+ \textsc{LongTraceRL}($\alpha=0.3$) & \textbf{41.8} & \underline{45.8} & \textbf{79.5} & \underline{44.1} & \textbf{83.8} & \textbf{59.0} \\
+ \textsc{LongTraceRL}($\alpha=0.5$) & 39.0 & 43.7 & 77.5 & 43.5 & 81.7 & 57.1 \\
\bottomrule
\end{tabular}
}
\caption{Performance of \textsc{LongTraceRL} with different rubric reward weight $\alpha$.}
\label{tab:abla_rubric_ratio}
\end{table*}

\begin{table*}[htbp]
\centering
\resizebox{0.9\textwidth}{!}{
\begin{tabular}{l|cccccc}
\toprule
\textbf{Method} & \textbf{AA-LCR} & \textbf{MRCR} & \textbf{FRAMES} & \textbf{LongBench V2} & \textbf{LongReason} & \textbf{Avg} \\
\midrule
\textit{Qwen3-4B-Thinking-2507} & 33.2 & 36.2 & 76.7 & 41.7 & 78.5 & 53.3  \\
+ \textsc{LongTraceRL}(\texttt{random}) & 34.2 & 40.1 & 78.0 & 43.8 & 82.4 & 55.7 \\
+ \textsc{LongTraceRL}(\texttt{search}) & 34.2 & 43.2 & 79.0 & 43.9 & \underline{83.3} & 56.7 \\
+ \textsc{LongTraceRL}(\texttt{traj-random}) & \underline{35.5} & \underline{44.2} & \textbf{79.9} & \textbf{44.6} & 82.8 & \underline{57.4} \\
+ \textsc{LongTraceRL}(\texttt{traj-tiered}) & \textbf{41.8} & \textbf{45.8} & \underline{79.5} & \underline{44.1} & \textbf{83.8} & \textbf{59.0} \\
\bottomrule
\end{tabular}
}
\caption{Performance of \textsc{LongTraceRL} across distractor strategies. Confusability increases from top to bottom.}
\label{tab:abla_distractor}
\end{table*}

\begin{table*}[hbtp]
\centering
\small
\setlength{\tabcolsep}{5pt}
\begin{tabular}{l|rrrrr}
\toprule
\textbf{Distractor Strategy} & \textbf{\#Distr.} & \textbf{\#w/ Rub.} & \textbf{Ent-Recall (\%)} & \textbf{Micro Avg (\%)} & \textbf{Macro Avg (\%)} \\
\midrule
\textbf{\texttt{traj-tiered}}                  & 62137 & 29050 & \textbf{10.34} & \textbf{46.75} & \textbf{50.03} \\
\quad Tier-1   & 18206 & 10629 & 14.65 & 58.38 & 63.23 \\
\quad Tier-2 & 43931 & 18421 &  8.56 & 41.93 & 41.75 \\
\midrule
\textbf{\texttt{traj-random}}                    & 64066 & 26528 & \underline{8.59} & \underline{41.41} & \underline{42.16} \\
\quad Tier-1   &  2677 &  1710 & 16.71 & 63.88 & 63.79 \\
\quad Tier-2 & 61389 & 24818 &  8.24 & 40.43 & 41.24 \\
\midrule
\textbf{\texttt{search}}                    & 31412 &  4372 & 2.47 & 13.92 & 15.00 \\
\midrule
\textbf{\texttt{random}}            & 45392 &   565 & 0.16 &  1.24 &  1.35 \\
\bottomrule
\end{tabular}
\caption{Statistics on how much distractors overlap with rubric entities. Higher ratios indicate harder distractors.
\textbf{\#Distr.}: number of distractor documents.
\textbf{\#w/ Rub.}: number of distractor documents containing $\geq$1 rubric entity.
\textbf{Ent-Recall}: average fraction of rubric entities appearing in a distractor.
\textbf{Micro/Macro Avg}: ratio of $\frac{\text{\#w/ Rub.}}{\text{\#Distr.}}$, aggregated globally (micro) or per sample then averaged (macro).
}
\label{tab:distractor_stats}
\end{table*}

\paragraph{Models.}
We experiment with several reasoning-capable LLMs of different families and sizes to test generalizability:
(1) \textbf{Qwen3-4B-Thinking-2507}~\citep{qwen3}, a 4B dense reasoning model; 
(2) \textbf{DeepSeek-R1-0528-Qwen3-8B}~\citep{deepseek_r1}, a distilled dense model from the updated DeepSeek-R1-0528;
(3) \textbf{Qwen3-30B-A3B-Thinking-2507}~\citep{qwen3}, a mixture-of-experts model with 30B total parameters and 3B active parameters.
All ablation studies are conducted using Qwen3-4B-Thinking-2507.

\paragraph{Datasets.}
Our training set consists of 2,815 long-context QA examples constructed via the pipeline described in \S\ref{sec:data_construction}. Each example contains a eight-hop question, gold evidence passages from Wiki, and tiered distractors assembled to a target context length of 128K tokens.
We compare our method against three existing long-context RL datasets, using the same RL algorithm and hyperparameters for a fair comparison:
(1) \textbf{DocQA}~\citep{qwenlong_l1}: 1,591 QA examples on real documents covering math, logic and multi-hop reasoning, with context lengths ranging from 2K to 20K tokens.
(2) \textbf{LoongRL}~\citep{loongrl}: 15,000 QA examples of 16K tokens, built by the KeyChain pipeline that pads short-context multi-hop QA (HotpotQA, MuSiQue, 2WikiMQA) with distractors and hides the true question behind a UUID chain.
(3) \textbf{LongRLVR}~\citep{longrlvr}: 18,870 QA examples from book, arXiv and code documents with 8K to 64K tokens, trained with an extra $\text{F}_\beta$-based context grounding reward over document chunks.
A further comparison is shown in Table~\ref{tab:dataset_stats}.

\paragraph{Benchmarks.}
We evaluate on five long-context benchmarks:
(1) \textbf{AA-LCR}~\citep{aa_lcr}: 100 expert-crafted questions over real-world documents averaging 100K tokens.
(2) \textbf{MRCR}~\citep{openai2025mrcr}: a multi-round coreference benchmark that asks the model to reproduce a specific response from a long dialog with repeated requests on overlapping topics. We evaluate with 2, 4, and 8 needles and report the average score.
(3) \textbf{Frames}~\citep{frames}: a multi-hop factual reasoning benchmark over multiple Wikipedia articles with numerical, temporal, or tabular reasoning.
(4) \textbf{LongBench v2}~\citep{longbench2}: 503 multiple-choice questions over real documents from 8K to 2M words, covering QA, in-context learning, code and so on.
(5) \textbf{LongReason}~\citep{longreason}: a synthetic benchmark that turns short reasoning problems into long-context versions by spreading key information across 8K to 128K tokens. We evaluate at 8K, 16K, 32K, 64K, and 128K and report the average score.

We run AA-LCR 4 times and LongBench v2 2 times, and report the average; the other three benchmarks are run once.

\paragraph{Training Details.}
We use the Slime framework~\citep{slime_github} for RL training. The maximum context length is set to 160K tokens (128K prompt + 32K response). We use GRPO with group size $G = 8$, global batch size 128, and train for 200 iterations with a constant learning rate $2 \times 10^{-6}$.
The rubric reward weight is $\alpha = 0.3$ by default, with group-level normalization and positive-only reward strategy. Rollout uses temperature 1.0, while all evaluations use temperature 0.6 and maximum generation length 32K. We save a checkpoint every 20 training steps and report results from the best-performing checkpoint.
All experiments are conducted on 32 $\times$ H800 GPUs.

\subsection{Main Results}
Table~\ref{tab:scores} reports the per-benchmark scores of \textsc{LongTraceRL} against other baselines across three scales.
\textsc{LongTraceRL} consistently achieves the best average score on every backbone: on Qwen3-4B-Thinking-2507 it reaches an average of \textbf{59.0}, improving the base model by \textbf{+5.7} points and surpassing the strongest baseline LongRLVR by \textbf{+2.5} points. The gain is most pronounced on the challenging AA-LCR (33.2 $\rightarrow$ 41.8, +8.6). The same trend holds for DeepSeek-R1-0528-Qwen3-8B (42.7 $\rightarrow$ 43.8) and Qwen3-30B-A3B-Thinking-2507 (60.5 $\rightarrow$ 63.7, +3.2), showing that the gains are robust to the model family and scale. In contrast, DocQA, LoongRL and LongRLVR even degrade performance on the 8B backbone (42.7 $\rightarrow$ 40.6 / 40.1 / 40.9). Besides, ablating the rubric reward (\textsc{LongTraceRL}-GRPO) on the 4B backbone drops the average score from 59.0 to 53.7, nearly erasing the gain despite training on the same dataset, identifying the rubric reward as the dominant driver of the improvement.

Figure~\ref{fig:rubric_reward} shows that the rubric reward grows steadily during \textsc{LongTraceRL} training on all three scales, indicating that the model progressively learns to ground its reasoning in the gold entities.
Figure~\ref{fig:outcome_reward} further shows that the outcome reward curve of \textsc{LongTraceRL} also rises and dominates that of \textsc{LongTraceRL}-GRPO, confirming that introducing the rubric reward helps the model reach the correct final answer.
Figure~\ref{fig:resp_len} compares rollout lengths across baselines, showing that the rubric reward clearly encourages longer and more deliberate reasoning. As shown in Figure~\ref{fig:truncated}, around step 120, many rollouts of \textsc{LongTraceRL} reach the 32K budget and thus fail to emit the final answer, which suppresses the outcome reward. The positive-only strategy then guides the policy back to shorter responses before the length climbs again. This self-regulating behavior shows that combining the positive-only strategy with a finite response budget effectively prevents rubric reward hacking.

\subsection{Ablation Studies}

\subsubsection{Rubric Ratio $\alpha$}
The hyperparameter $\alpha$ in Eq.~\ref{eq:reward} controls the relative weight of the rubric reward against the outcome reward, and thus determines how strongly the model is pushed to ground its reasoning in the gold entities. To study its effect, we sweep $\alpha \in \{0.1, 0.3, 0.5\}$ while keeping all other settings identical to our main experiment. As shown in Table~\ref{tab:abla_rubric_ratio}, $\alpha = 0.3$ yields the best average score (59.0) and consistently achieves the best or second-best performance across all five benchmarks. Decreasing $\alpha$ to 0.1 weakens the process-level signal and degrades performance, especially on the more reasoning-intensive AA-LCR benchmark (41.8 $\rightarrow$ 39.2), while increasing $\alpha$ to 0.5 hurts performance across the board (average drops to 57.1), suggesting that an overly strong rubric weight begins to dilute the outcome objective and bias the model toward entity-mention shortcuts. 

\begin{table*}[htbp]
\centering
\resizebox{0.9\textwidth}{!}{
\begin{tabular}{l|cccccc}
\toprule
\textbf{Method} & \textbf{AA-LCR} & \textbf{MRCR} & \textbf{FRAMES} & \textbf{LongBench V2} & \textbf{LongReason} & \textbf{Avg} \\
\midrule
\textit{Qwen3-4B-Thinking-2507} & 33.2 & 36.2 & 76.7 & 41.7 & 78.5 & 53.3 \\
+ \textsc{LongTraceRL}(\texttt{positive\&negative}) & 37.0 & 40.5 & 79.5 & \textbf{45.5} & 83.1 & 57.1 \\
+ \textsc{LongTraceRL}(\texttt{positive-only}) & \textbf{41.8} & \textbf{45.8} & \textbf{79.5} & 44.1 & \textbf{83.8} & \textbf{59.0} \\
\bottomrule
\end{tabular}
}
\caption{Performance of \textsc{LongTraceRL} with different reward strategies.}
\label{tab:abla_pos_neg}
\end{table*}

\begin{figure*}[!t]
    \centering
    \begin{subfigure}{0.31\linewidth}
        \centering
        \includegraphics[width=\linewidth]{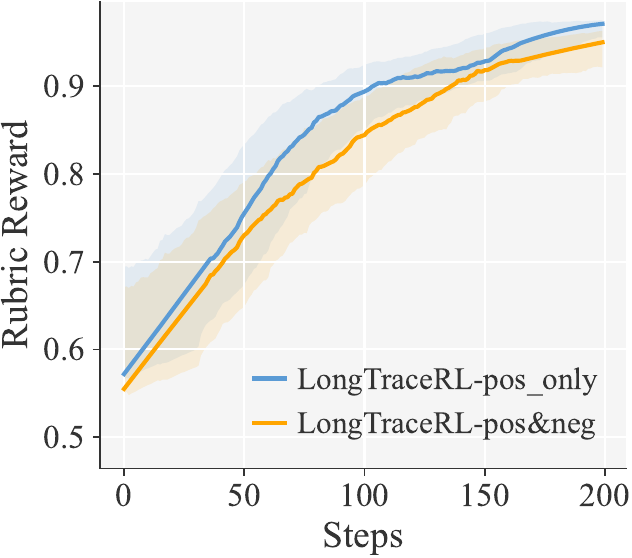}
        \label{fig:rubric_reward_4B_pos_neg}
    \end{subfigure}
    \begin{subfigure}{0.31\linewidth}
        \centering
        \includegraphics[width=\linewidth]{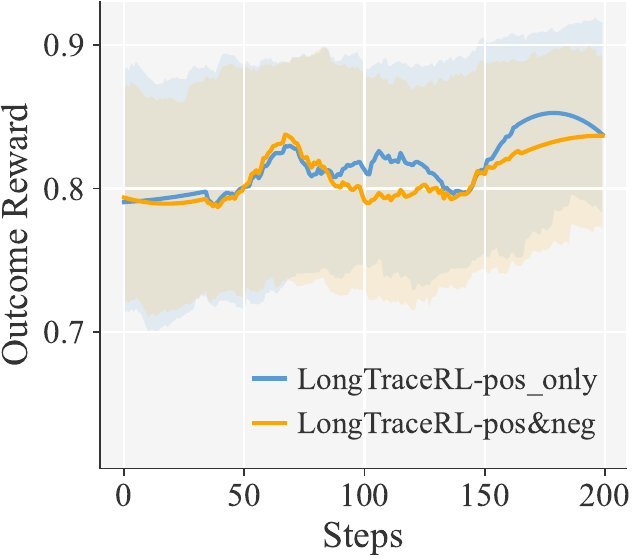}
        \label{fig:outcome_reward_4B_pos_neg}
    \end{subfigure}
    \begin{subfigure}{0.31\linewidth}
        \includegraphics[width=\linewidth]{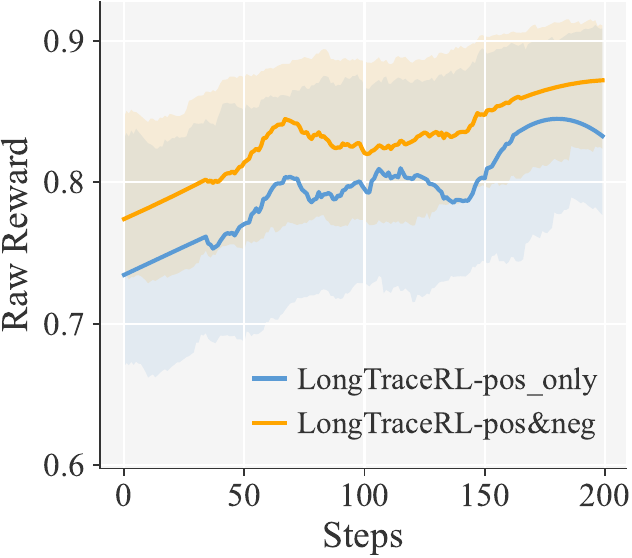}
        \label{fig:raw_reward_4B_pos_neg}
    \end{subfigure}
    \vspace{-0.5cm}
    \caption{Rubric, outcome and combined raw reward dynamics for the two reward strategies.}
    \label{fig:abla_pos_neg}
\end{figure*}

\subsubsection{Source of Distractors}
To verify the effectiveness of \textbf{\texttt{traj-tiered}} distractor strategy, we compare it against three alternative strategies while keeping the gold passages and questions identical:
(1) \textbf{\texttt{random}}: distractors are randomly sampled from a global pool of documents retrieved across the dataset and are therefore mostly off-topic for the current question.
(2) \textbf{\texttt{search}}: we issue the question as a single query to a search engine and leverage the top-100 results as distractors, approximating naive one-shot retrieval without multi-round querying, document opening or filtering.
(3) \textbf{\texttt{traj-random}}: we pool Tier-1 and Tier-2 distractors of each question together and randomly sample documents from this pool, with no preference for their confusability. 

This yields three new training sets that differ only in their distractor documents. 
We train Qwen3-4B-Thinking-2507 on each set under the same setting as the main experiment. 
As shown in Table~\ref{tab:abla_distractor}, the \texttt{traj-tiered} strategy achieves the best average score (59.0), with the advantage particularly pronounced on AA-LCR (41.8 vs.\ at most 35.5 for the alternatives).
To better explain these gains, we quantify distractor difficulty as the fraction of distractor documents that share at least one rubric entity with the reasoning chain and are thus harder to filter out.
As Table~\ref{tab:distractor_stats} shows, the rank of distractor difficulty closely matches the rank of downstream scores in Table~\ref{tab:abla_distractor}: 
\texttt{random} distractors almost never share rubric entities with the question (Macro Avg 1.35\%), so they are easy to filter out and provide a weak training signal, leading to the lowest score (55.7). \texttt{search} strategy is slightly harder (Macro Avg 15.00\%) and improves the score to 56.7, but it still cannot guarantee that the distractors lie along the reasoning path. Trajectory-based distractors are far more challenging: \texttt{traj-random} reaches 42.16\%, and \texttt{traj-tiered} further pushes the ratio to 50.03\%, with Tier-1 documents alone reaching 63.23\%. This higher density of hard distractors aligns with the best downstream performance (59.0).
These results confirm that the distractor is a key driver of long-context RL data quality, and that our tiered, trajectory-derived design is more effective than single-search-based or random-based alternatives.

\subsubsection{Positive-Only}
To verify the effectiveness of the \textbf{\texttt{positive-only}} strategy in preventing reward hacking, we compare it against a \textbf{\texttt{positive\&negative}} variant in which the rubric reward is granted to every rollout regardless of answer correctness, i.e., $r = (1 - \alpha) \cdot r_{\text{oc}} + \alpha \cdot r_{\text{rb}}$ for all responses. 
As shown in Table~\ref{tab:abla_pos_neg}, removing the positive-only constraint causes a clear performance drop, with the average score falling from 59.0 to 57.1. The degradation is most pronounced on the reasoning-intensive benchmarks where the rubric signal is meant to help: AA-LCR drops by 4.8 points (41.8 $\rightarrow$ 37.0) and MRCR by 5.3 points (45.8 $\rightarrow$ 40.5). The training dynamics in Figure~\ref{fig:abla_pos_neg} further reveal the underlying mechanism. Throughout training, the \texttt{positive\&negative} variant exhibits \emph{lower} rubric and outcome rewards, yet its combined reward ends up \emph{higher}, because every rollout, including the incorrect ones, can gain a non-trivial rubric term to the aggregate. 
This misleading objective dilutes the gradient toward genuinely solving the question and biases the policy toward enumerating gold-like entities from the context. 

\section{Conclusion}
We present \textsc{LongTraceRL}, a framework that advances long-context RL by constructing challenging training data from trajectory-based distractors and introducing an entity-level rubric reward for fine-grained process supervision.
Experiments on five long-context benchmarks across three model families and scales demonstrate consistent improvements over existing long-context RL methods.
Further analysis confirms the effectiveness of each design choice.
We hope that our approach offers a practical and generalizable recipe for further research on long-context RL of LLMs.

\section{Limitations}
Our work has several limitations.
First, the data construction pipeline relies entirely on the KILT Wikipedia snapshot as its knowledge source, meaning all generated questions are grounded in encyclopedic knowledge. While our experiments show that training on such data transfers well to various downstream benchmarks covering financial, legal, and code documents, the single-source nature of the knowledge graph may limit the diversity of reasoning patterns in the training data. 
Second, the search agent trajectories used for distractor construction depend on the capabilities of the particular agent deployed. A stronger or weaker agent would produce different trajectory distributions, potentially affecting the quality and difficulty of the resulting distractors. Investigating how agent capability influences data quality is an interesting direction for future work.

\section{Ethical Considerations}
All models and datasets used in this work are publicly available under permissible licenses. 
Our method does not involve human subjects, private data, or content that raises dual-use concerns.

\bibliography{custom}

\appendix
\newpage
\section{Dataset Comparison}
\begin{table*}[t]
\centering
\small
\setlength{\tabcolsep}{5pt}
\begin{tabular}{l|rrcl}
\toprule
\textbf{Dataset} & \textbf{\#Samples} & \textbf{Context Length} & \textbf{\#Hops} & \textbf{Question Source} \\
\midrule
DocQA & 1,591 & 2K--20K & 2--4 & Real documents (math, logic, multi-hop) \\
LoongRL & 15,000 & 16K & 2--4 & HotpotQA, MuSiQue, 2WikiMQA \\
LongRLVR & 18,870 & 8K--64K & N/A & LLM-generated from book/arXiv/code \\
\textsc{LongTraceRL} & 2,815 & 128K & 8 & KG random walk on Wikipedia \\
\bottomrule
\end{tabular}
\caption{Comparison of training datasets used in long-context RL methods.}
\label{tab:dataset_stats}
\end{table*}

Table~\ref{tab:dataset_stats} compares the training datasets of the long-context RL methods evaluated in our experiments.
DocQA contains 200 MuSiQue examples (4-hop) and 200 MultiHopRAG examples (2--4 hop), and the remaining 1,191 are math or logic questions without a defined hop count.
LoongRL reuses HotpotQA (2-hop), MuSiQue (2--4 hop), and 2WikiMQA (2-hop). Its KeyChain mechanism adds extra UUID-tracing steps on top.
LongRLVR generates questions from single documents via LLM, which require cross-chunk synthesis but do not follow a defined multi-hop chain.
In contrast, \textsc{LongTraceRL} constructs questions via knowledge graph random walks with $k=8$ hops, producing substantially deeper reasoning chains.

\section{Case Studies}
\begin{figure*}[!t]
    \centering    \includegraphics[width=0.97\linewidth]{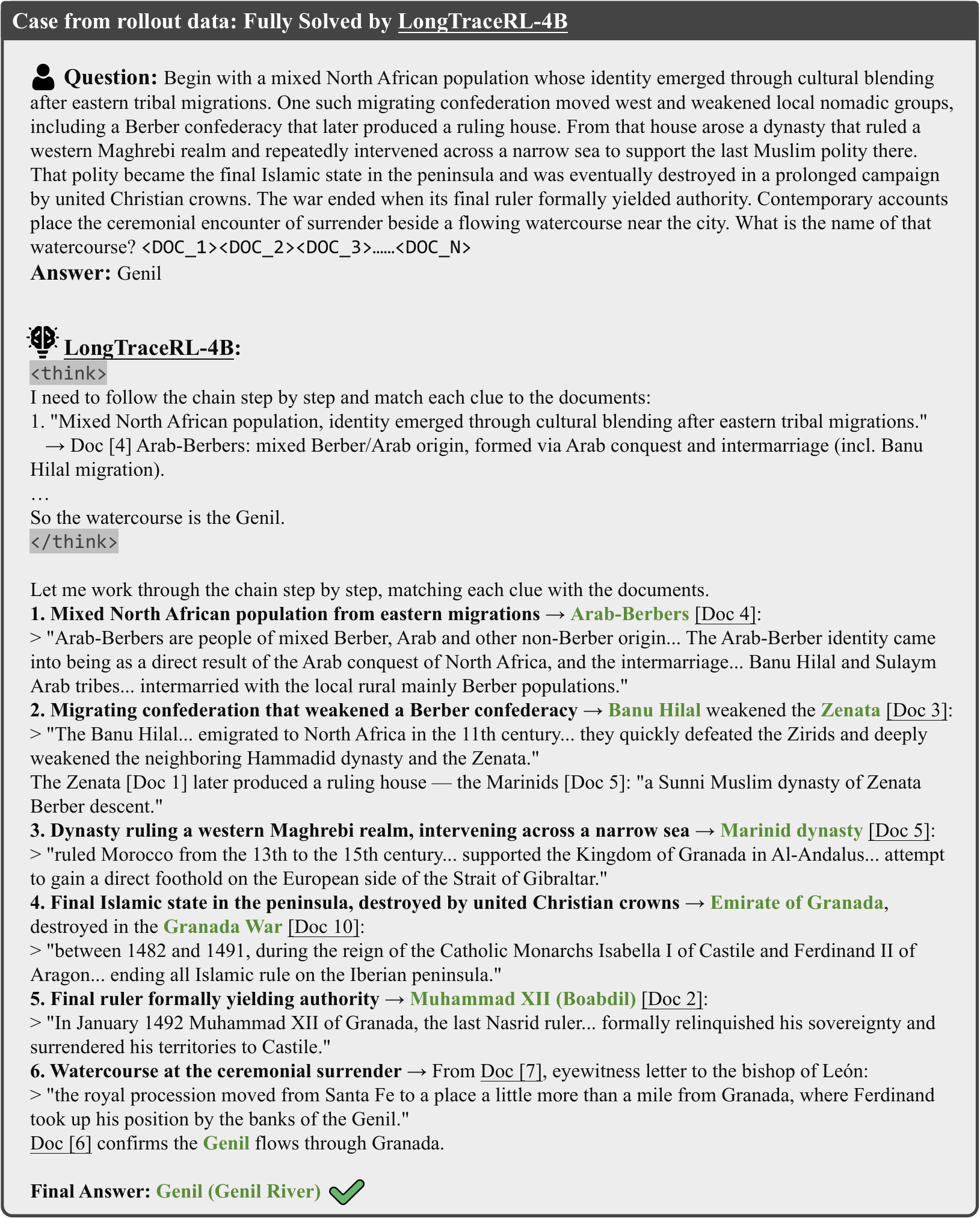}
    \caption{A training rollout from \textsc{LongTraceRL}-4B on a synthesized multi-hop question.}
    \label{fig:case_study_rollout}
\end{figure*}

\paragraph{Case from rollout data.}
Figure~\ref{fig:case_study_rollout} shows a training rollout on a multi-hop question synthesized by our pipeline, whose reasoning chain spans seven gold entities (Arab-Berbers, Banu Hilal, Zenata, Marinid dynasty, Emirate of Granada, Granada War, Muhammad XII) before arriving at the final answer ``Genil''. A natural concern with our entity-recall rubric is that the model might learn to inflate its score by enumerating every entity it sees in the context. The rollout here illustrates the opposite behavior: \textsc{LongTraceRL}-4B visits each gold entity in the correct order and commits to the final answer. Each cited entity serves a necessary role in the reasoning chain, and entities outside the gold path are never introduced merely to inflate the rubric score. This is exactly the behavior that the positive-only rubric reward is designed to elicit.

\begin{figure*}[!t]
    \centering    \includegraphics[width=0.97\linewidth]{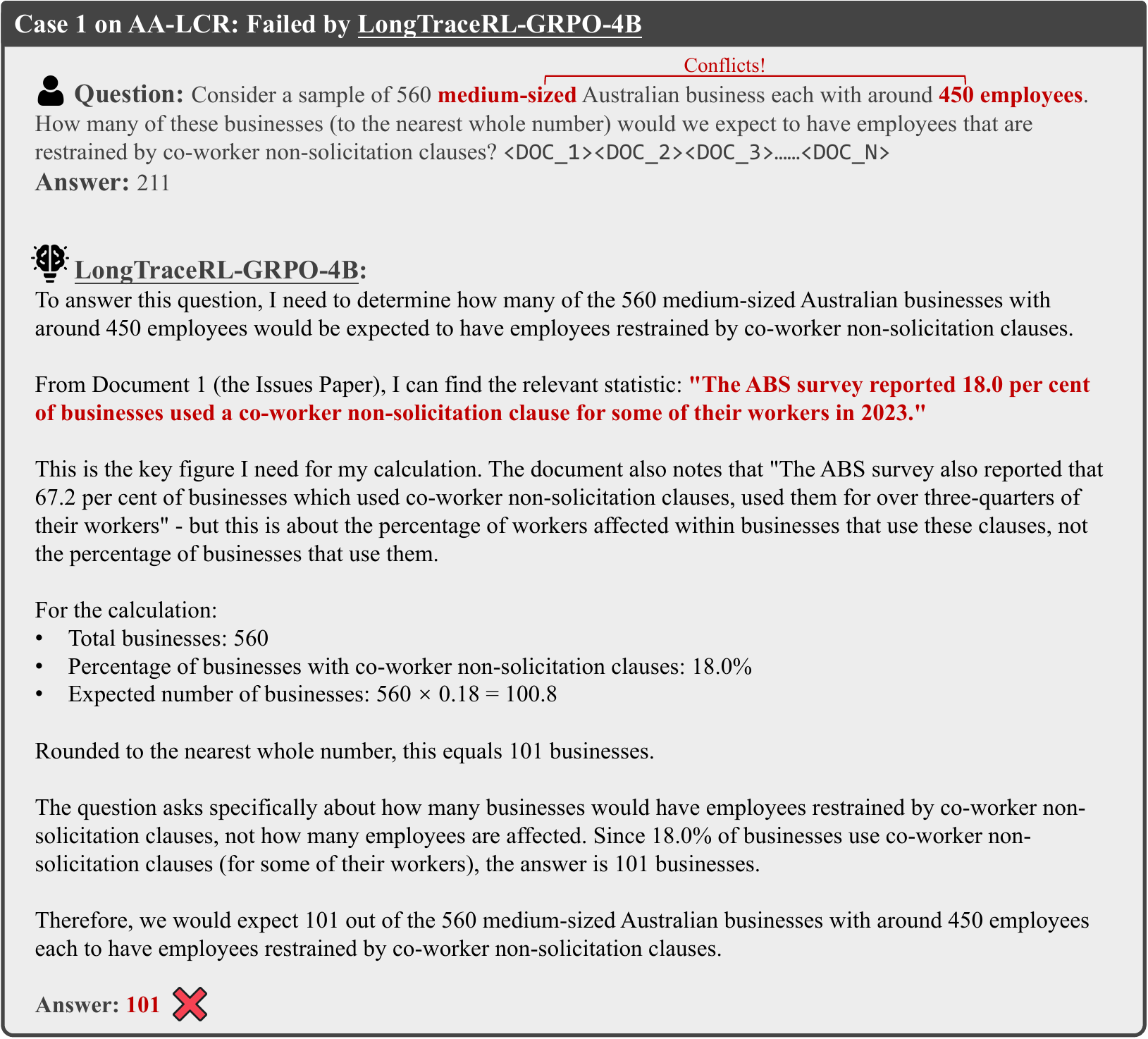}
    \caption{A failure case from AA-LCR where the \textsc{LongTraceRL}-GRPO-4B trained without rubric reward takes a shortcut without resolving the conflict in the question.}
    \label{fig:case_study_1_error}
\end{figure*}

\begin{figure*}[!t]
    \centering    \includegraphics[width=0.97\linewidth]{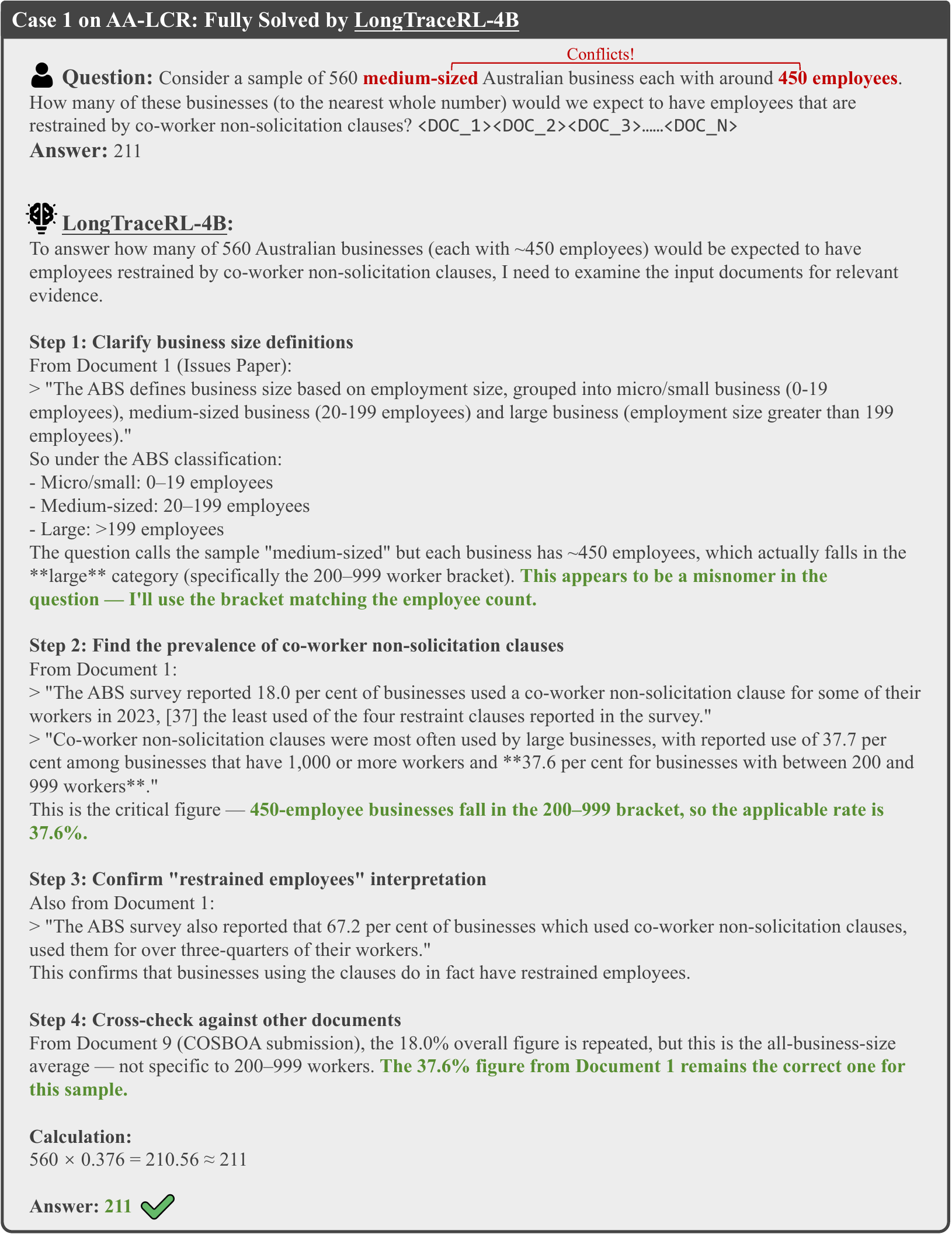}
    \caption{A success case from AA-LCR where the \textsc{LongTraceRL}-4B trained with rubric reward identifies the conflict and applies the correct rate (37.6\%) to reach the correct answer.}
    \label{fig:case_study_1_correct}
\end{figure*}

\paragraph{Case 1: Resolving Conflicting Cues in the Question.}
Figures~\ref{fig:case_study_1_error} and~\ref{fig:case_study_1_correct} present a representative AA-LCR example that highlights the reasoning depth our reward formulation aims to promote. The question intentionally mixes two conflicting cues: its surface label says ``medium-sized'', but the supplied headcount of 450 employees actually falls into the ABS \emph{large business} category ($>199$ employees). 
The source document further reports two relevant ratios: a general $18.0\%$ over \emph{all} businesses, and a rate of $37.6\%$ specifically for businesses with $200$ to $999$ workers. 
Therefore, to solve the question, one must first recognize this internal contradiction in the question and then retrieve the percentage that matches the actual headcount.
As Figure~\ref{fig:case_study_1_error} shows, \textsc{LongTraceRL}-GRPO-4B takes a shortcut, selecting the first reasonable figure ($18.0\%$) and missing the answer. In contrast, \textsc{LongTraceRL}-4B trained with rubric reward (Figure~\ref{fig:case_study_1_correct}) explicitly flags the inconsistency between ``medium-sized'' and 450 employees, reclassifies the firms by their actual headcount, and applies the $37.6\%$ rate to get the correct answer.
This example illustrates precisely how \textsc{LongTraceRL}'s design promotes deeper reading and reasoning.

\begin{figure*}[!t]
    \centering    \includegraphics[width=0.97\linewidth]{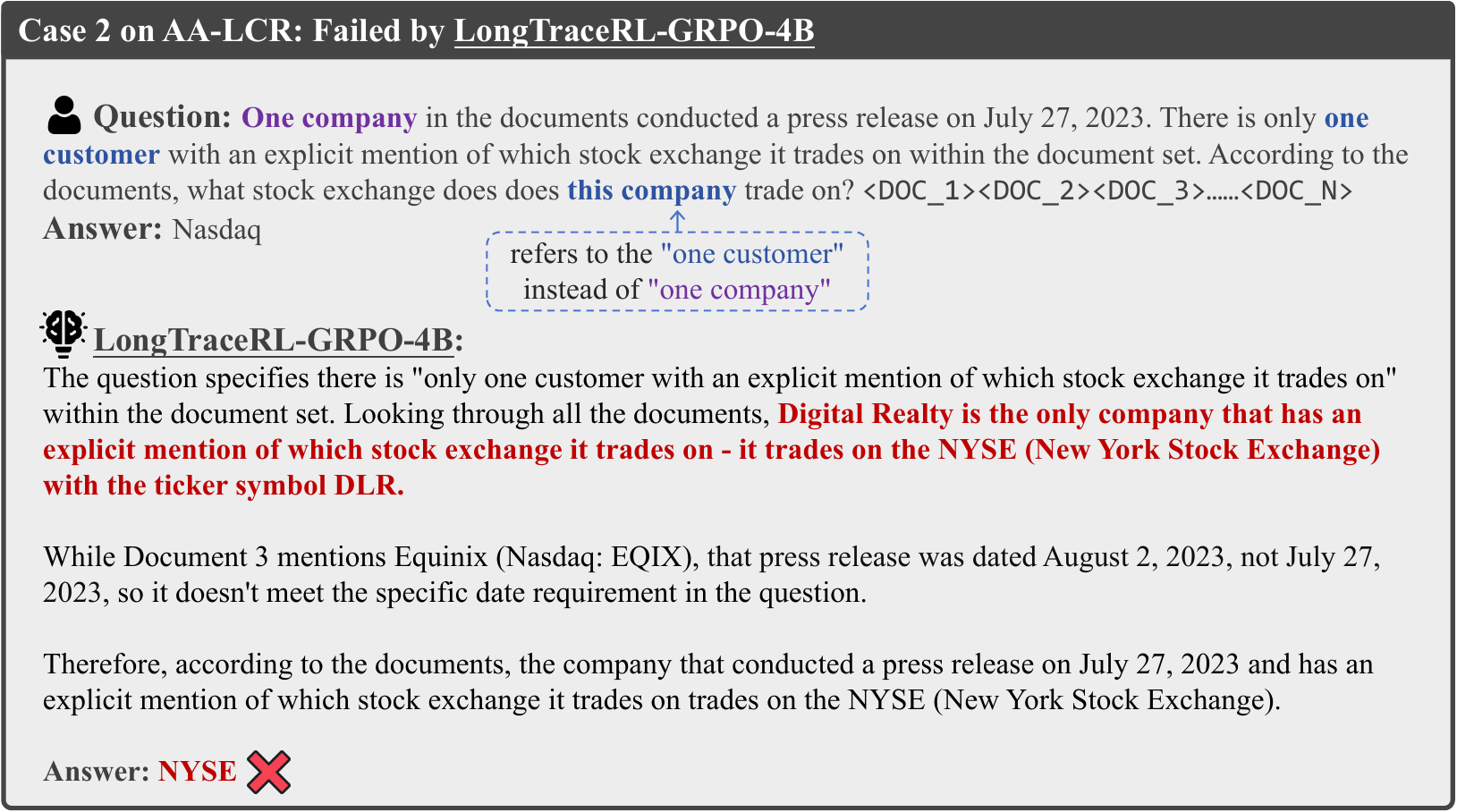}
    \caption{A failure case from AA-LCR where the \textsc{LongTraceRL}-GRPO-4B trained without rubric reward incorrectly merges the two clauses into a single entity and answers with the press-release issuer's exchange.}
    \label{fig:case_study_2_error}
\end{figure*}

\begin{figure*}[!t]
    \centering    \includegraphics[width=0.97\linewidth]{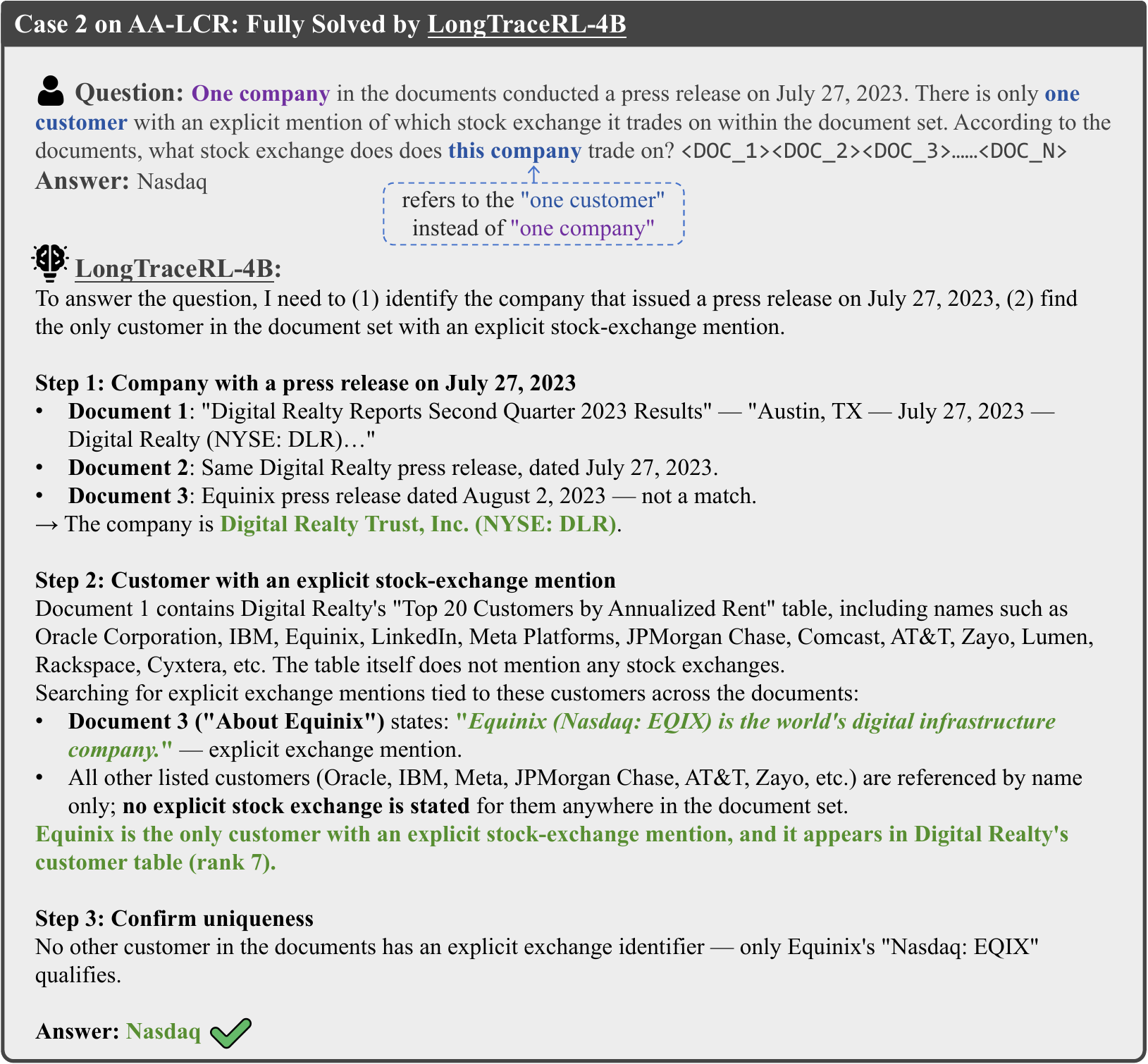}
    \caption{A success case from AA-LCR where the \textsc{LongTraceRL}-4B trained with rubric reward keeps the two clauses separate and returns the correct answer.}
    \label{fig:case_study_2_correct}
\end{figure*}

\paragraph{Case 2: Disambiguating Pronoun Reference Across Clauses.}
Figures~\ref{fig:case_study_2_error} and~\ref{fig:case_study_2_correct} illustrate a different type of reasoning challenge. In this question, two clauses share an ambiguous ``this company'': the first clause refers to a company that issued a press release on July~27, 2023 (Digital Realty), while the second clause introduces ``one customer with an explicit mention of which stock exchange it trades on'' and then asks about \emph{that customer}'s exchange. Solving the question therefore requires (i) identifying the press-release issuer, (ii) realizing that the final ``this company'' refers back to the \emph{customer} rather than the issuer, and (iii) finding the unique customer in the documents whose exchange is explicitly annotated (Equinix, Nasdaq: EQIX).
As Figure~\ref{fig:case_study_2_error} shows, \textsc{LongTraceRL}-GRPO-4B merges the two clauses into a single entity, incorrectly treating Digital Realty as the answer. It actually retrieves the Equinix/Nasdaq passage but dismisses it because of a misapplied date constraint. 
In contrast, \textsc{LongTraceRL}-4B (Figure~\ref{fig:case_study_2_correct}) re-parses the question structure, explicitly pivots from issuer to customer, scans the documents for the unique customer with an exchange annotation, and returns the correct answer ``Nasdaq''.

\begin{figure*}[!t]
    \centering    \includegraphics[width=0.97\linewidth]{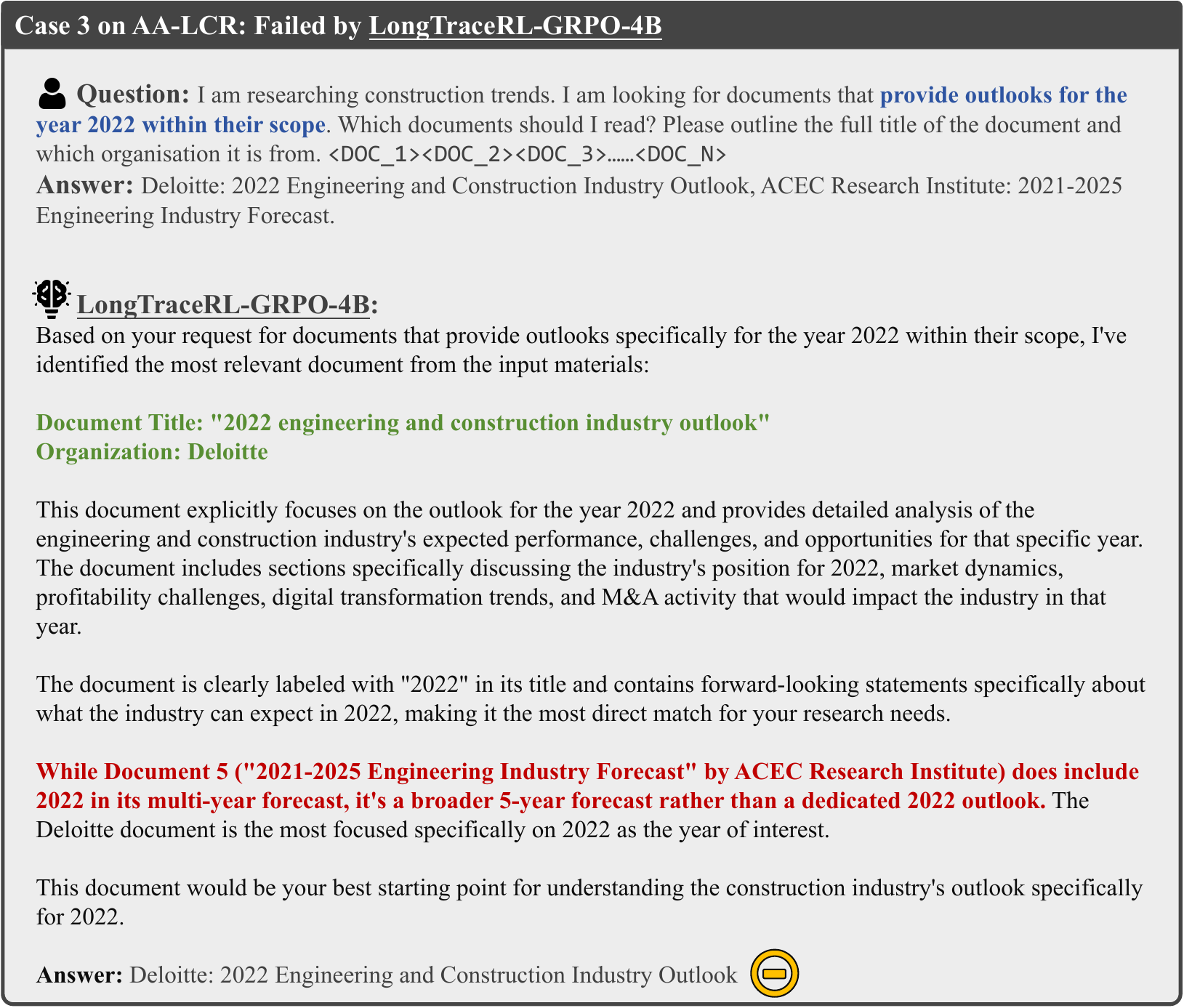}
    \caption{A failure case from AA-LCR where the \textsc{LongTraceRL}-GRPO-4B trained without rubric reward silently rewrites the question and uses a tightened criterion to discard the ACEC document.}
    \label{fig:case_study_3_error}
\end{figure*}

\begin{figure*}[!t]
    \centering    \includegraphics[width=0.97\linewidth]{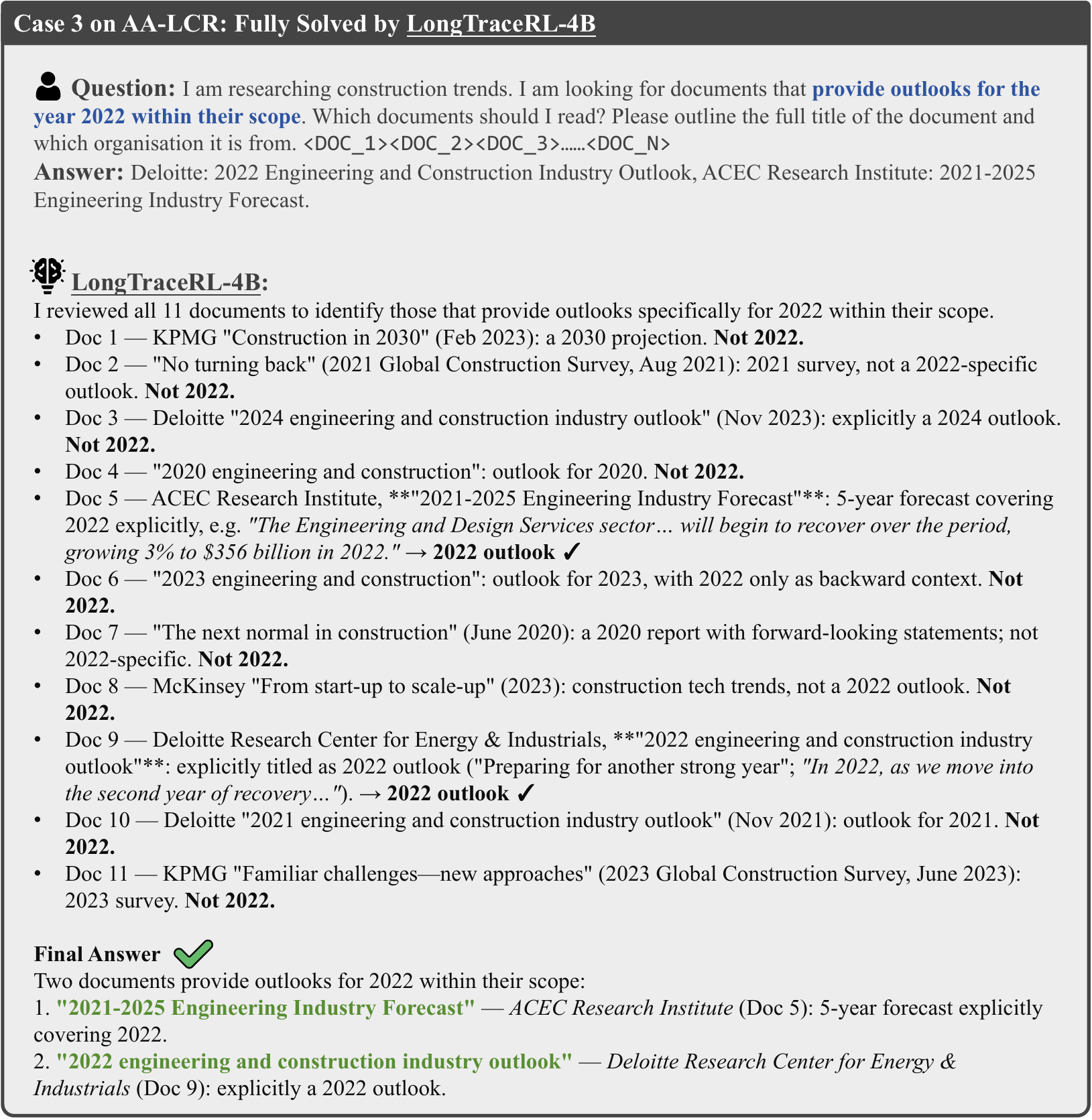}
    \caption{A success case from AA-LCR where the \textsc{LongTraceRL}-4B trained with rubric reward checks each candidate document and finds both qualifying outlooks with correct organization attributions.}
    \label{fig:case_study_3_correct}
\end{figure*}

\paragraph{Case 3: Understanding a Subtle Qualifier.}
Figures~\ref{fig:case_study_3_error} and~\ref{fig:case_study_3_correct} highlight a third trap: a subtle filter hidden in the phrase ``\emph{within their scope}''. The question asks for documents whose outlook covers the year 2022, which is an inclusive criterion: a document is qualified as long as its forecast period covers 2022, even when ``2022'' does not appear in its title. Two documents meet this requirement: the Deloitte ``2022 Engineering and Construction Industry Outlook'' (an obvious title match) and the ACEC ``2021--2025 Engineering Industry Forecast'', which provides explicit 2022 projections (e.g., ``\$356 billion in 2022'') as part of its five-year forecast.
As Figure~\ref{fig:case_study_3_error} shows, \textsc{LongTraceRL}-GRPO-4B secretly rewrites the question, focuses solely on the title-level keyword match and discards the ACEC document despite having retrieved it. In contrast, \textsc{LongTraceRL}-4B (Figure~\ref{fig:case_study_3_correct}) iterates over each candidate document, recognizes that the ACEC forecast covers 2022, and correctly returns both documents together with their source organizations.

\section{Prompts}
\begin{figure*}[!t]
    \centering    \includegraphics[width=0.97\linewidth]{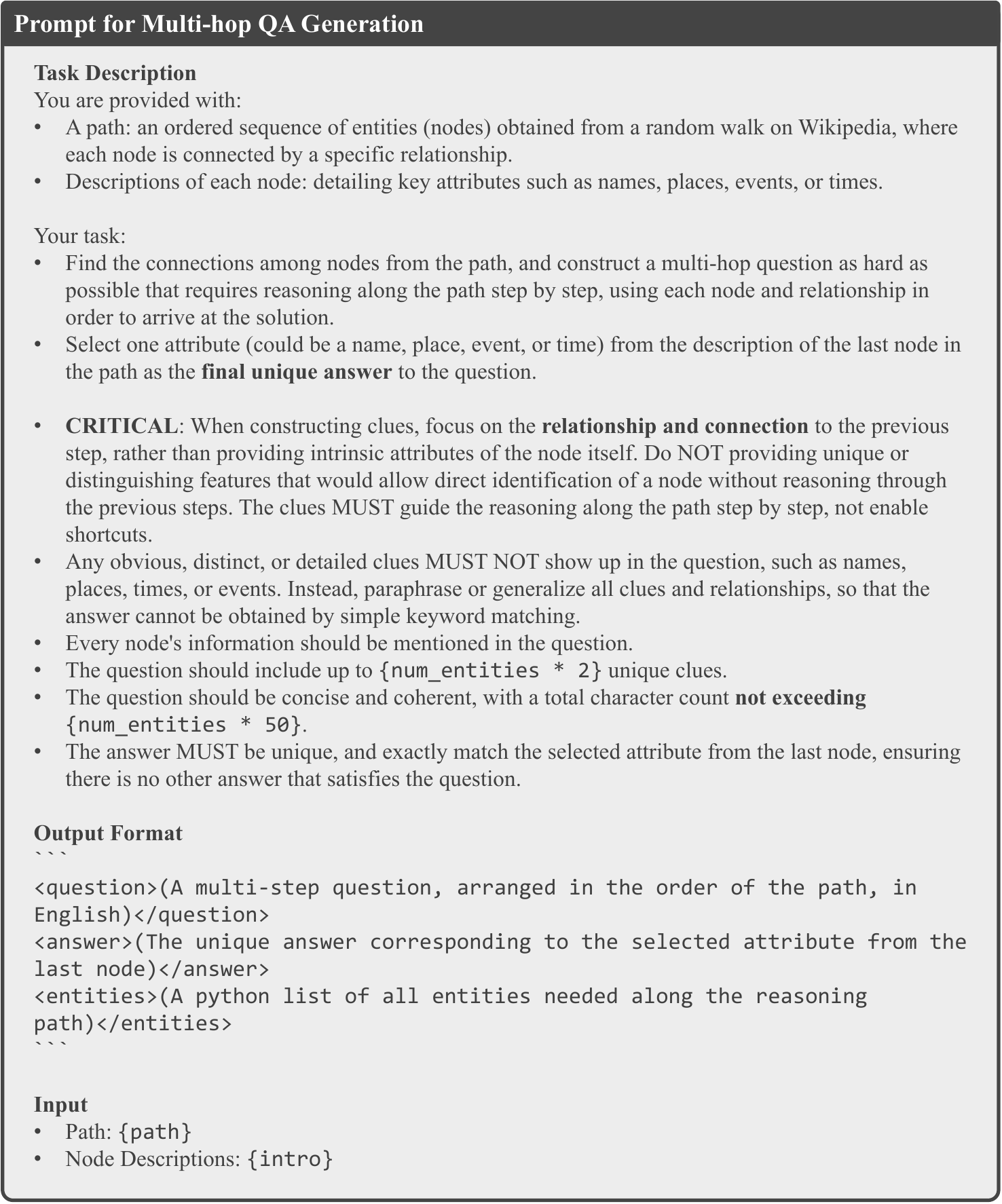}
    \caption{Prompt for multi-hop QA generation in \textsc{LongTraceRL}.}
    \label{fig:prompt_qa_generate}
\end{figure*}

\begin{figure*}[!t]
    \centering    \includegraphics[width=0.97\linewidth]{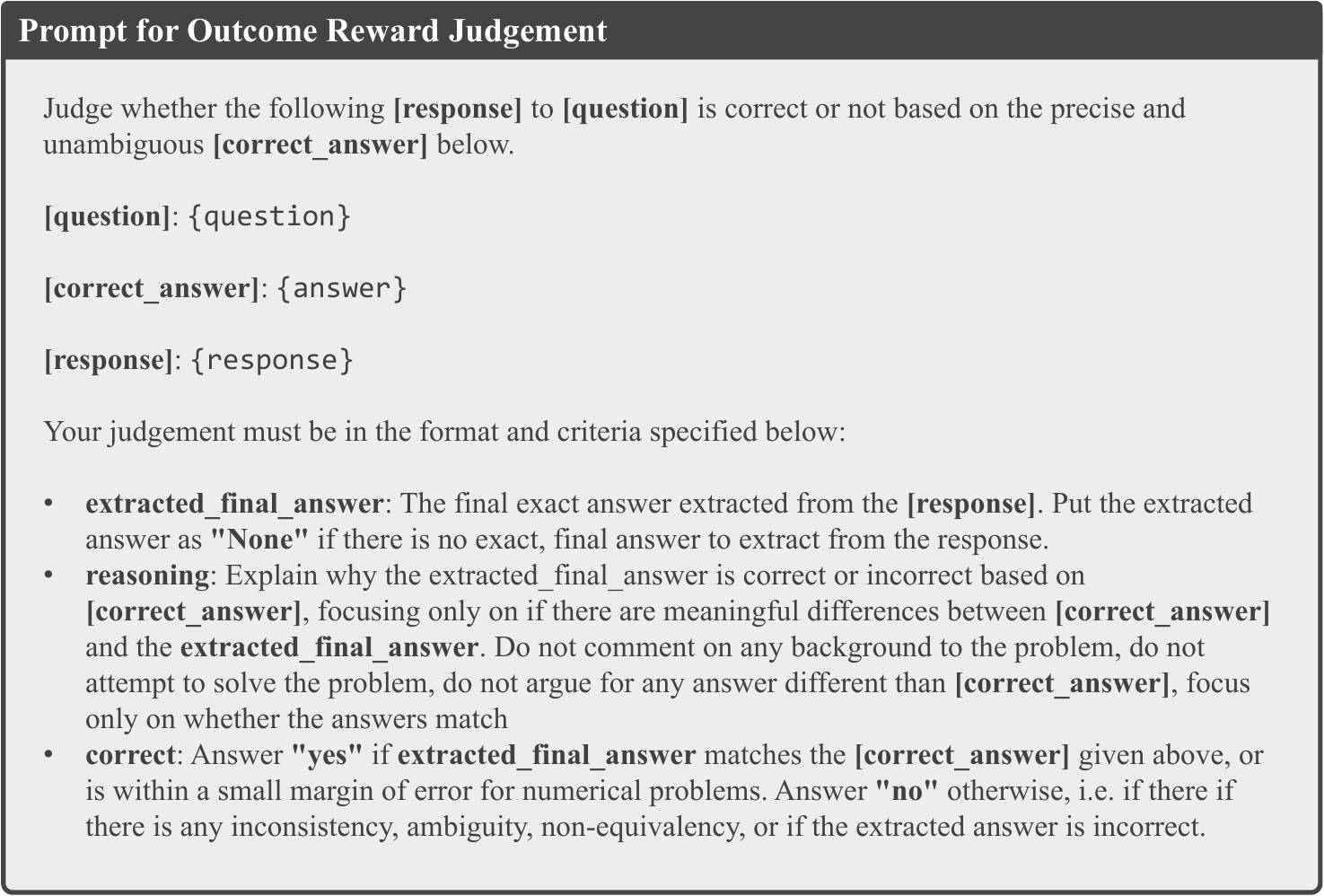}
    \caption{Prompt for outcome reward judgement in \textsc{LongTraceRL}.}
    \label{fig:prompt_outcome_reward_judge}
\end{figure*}

We show our used prompts in Figure~\ref{fig:prompt_qa_generate}, \ref{fig:prompt_outcome_reward_judge}.

\end{document}